# HANDWRITTEN CHARACTER RECOGNITION IN MALAYALAM SCRIPTS – A REVIEW

Anitha Mary M.O. Chacko and Dhanya P.M.

Department of Computer Science & Engineering, Rajagiri School of Engineering & Technology, Kochi, India

## ABSTRACT

*Handwritten character recognition is one of the most challenging and ongoing areas of research in the field of pattern recognition. HCR research is matured for foreign languages like Chinese and Japanese but the problem is much more complex for Indian languages. The problem becomes even more complicated for South Indian languages due to its large character set and the presence of vowels modifiers and compound characters. This paper provides an overview of important contributions and advances in offline as well as online handwritten character recognition of Malayalam scripts.*

## 1. INTRODUCTION

Automatic recognition of handwritten text has been a frontier area of research for the past few decades. It has wide range of applications in the domain of postal automation, automatic number plate recognition, preservation of handwritten historical documents, bank check processing, reading aid for the blind people etc. Handwritten character recognition is traditionally divided into online and offline recognition. In online recognition, the data is captured as a user writes on a special digitizer or PDA with stylus. In the offline case, the data is obtained by a scanner after the writing process is over. In online recognition, temporal and spatial information about each stroke is available whereas no such information is present in offline recognition making the recognition process even more complex.

Even though HCR systems are well advanced in foreign language scripts like Chinese and Japanese [1], only few works exist in Indian scripts especially in the South Indian scripts. This is mainly due to its large character set, high degree of similarity between these characters and the presence of compound characters in these scripts. Also, variations in the writing styles of different people make the recognition process more difficult. Among the Indian languages, most of the work has been reported in Devanagari [30] and Bangla. The research on South Indian scripts namely Malayalam, Tamil, Telungu and Kannada have gained much popularity recently as many agencies like the Ministry of Communications and Information Technology and Government of India are providing aid and financial support for many of these projects.

The outline of this paper is as follows: An overview of a general character recognition system is given in Section 2. Section 3 describes the features of Malayalam script. Some of the major works done in offline and online Malayalam character recognition domain are explained in Section 4 and 5 respectively. The conclusion and future works are presented in Section 6.

## 2. Components Of A General Character Recognition System

The major steps involved in the handwritten character recognition are :
- Data Acquisition
- Preprocessing
- Feature Extraction
- Segmentation
- Classification
- Post Processing

The architecture of a general handwritten character recognition system is illustrated in figure 1.

### 2.1. Data Acquisition

This is the stage in which data are collected as part of the recognition process. The data may be captured online while the user is writing on a digitizer or PDA. In the offline case, the data is obtained by scanning the image after the writing process is over.

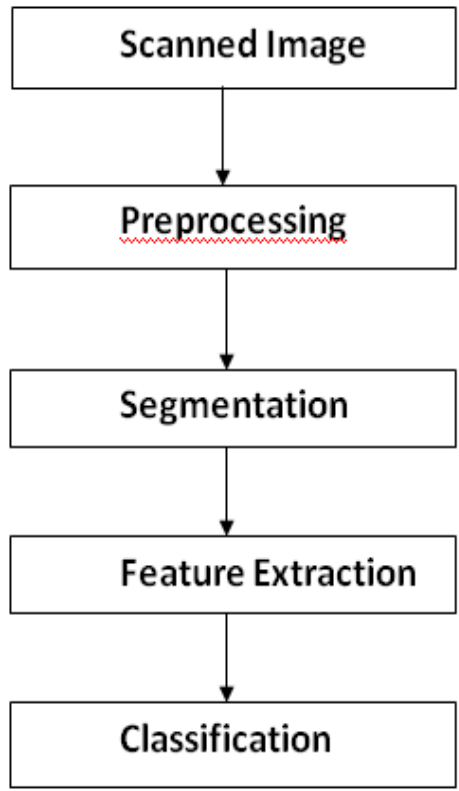

Figure 1 : Architecure of a general handwritten character recognition system

## 2.2. Preprocessing

### 2.2.1. Offline Documents

Preprocessing is an important step in character recognition process because of the variations in the writing style among different users and the existence of huge amount of noise in the images after scanning. The preprocessing steps involved in the offline case are:

#### 2.2.1.1. Binarization

Binarization is the process of converting grayscale images to binary images. It is done in order to identify the objects of interest from the image. It separates the foreground pixels from the background pixels. Figure 2 shows a sample scanned image and its corresponding binarized image. Binarization process is done by local or global thresholding. Local thresholding methods are based on applying different threshold values to different regions of the image. Niblack's method [2] and Sauvola's method [3] are two well known techniques for local thresholding. Global thresholding methods apply one threshold value to the entire image. Otsu's algorithm [4] is a commonly used approach for global thresholding.

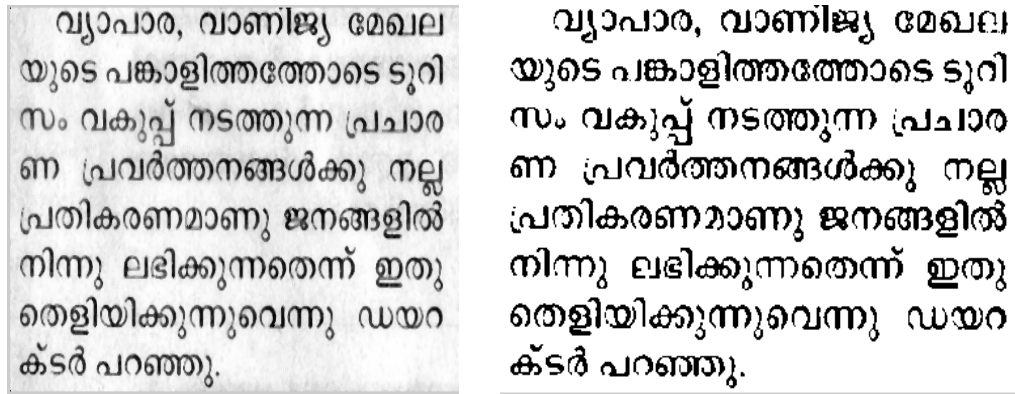

Figure 2 : a) Scanned Image          b) Binarized Image

#### 2.2.1.2. Noise Removal

A large amount of noise may occur in the image obtained after scanning. This may be due to the poor quality of the scanner or the use of degraded documents. Gaussian noise & Salt and Pepper noise are two such common noises. Filtering techniques such as linear filtering, median filtering or adaptive filtering can be used to remove noise to a certain extent. The main objective of this phase is to remove as much noise as possible while retaining the original signal.

#### 2.2.1.3. Skeletonization

Skeletonization or thinning is a morphological operation in which a single pixel wide representation of an image is obtained without changing its connectivity. The purpose of thinning is to reduce the image components so that they contain only essential information. The commonly used thinning algorithms are the Classical Hilditch algorithm, Zhang-Suen algorithm, Stentiford thinning algorithm. A survey of thinning methodologies is presented in [5].

**2.2.1.4. Skew Correction**

During document scanning, skew is introduced in the image. Skew angle is the angle that the text lines of the image make with the horizontal direction. The aim of the skew detection operation is to align an image before processing. Some of the major techniques available for skew estimation are projection profile method and Hough transform method. The different skew correction methods are described in [6]. The image is then rotated based on the detected skew angle.

**2.2.1.5. Normalization**

Normalization is the process of converting the image into a standard size. Bilinear and Bicubic interpolation techniques can be used for size normalization. Bilinear Interpolation determines the value of a new pixel based on a weighted average of the four pixels in the nearest 2 x 2 neighbourhood to the reference pixel in the original image. In the Bicubic Interpolation, new pixel value is a bicubic function using the weighted average of the 16 pixels in the nearest 4 x 4 neighbourhood of the reference pixel in the original image.

**2.2.2. Online Documents**

In online handwriting, each stroke is represented as a sequence of points taken from a pen down position to a pen-up position. The sequences of preprocessing steps commonly done for the online recognition are:

**2.2.2.1. Dehooking**

Hooks occur at the beginning and the end of character stroke and are generated by the pen-down and pen up movements. Dehooking is the process of eliminating such unwanted strokes that appear due to inaccuracies in pen down position. Dehooking algorithms are applied to remove hooks. Here, strokes are detected by comparing the number of points with a threshold value. If the value is greater than the threshold value, the mark is retained or it is removed otherwise.

**2.2.2.2. Duplicate Point Elimination**

Duplicate points are removed before feature extraction as these extra points do not contain any additional information. In this process, neighboring points having identical x-y coordinates are removed in this process.

**2.2.2.3. Smoothing**

Smoothing operation reduces variations that can occur due to fast writing or trembling of hand during the writing process. This operation is done in order to remove noise present in the acquired data. It removes unwanted cusps and self intersection using a linear filter. To perform smoothing, the current point is replaced by the mean value of its neighboring points.

**2.2.2.4. Size Normalization**

Normalization is the process of converting all characters to the predefined height and width without altering the aspect ratio. This is done by scaling transformation which converts all characters to the predefined constant width and height. This ensures that size of handwritten characters makes no difference in recognition. The size variations among handwritten characters are removed by this operation.

**2.2.2.5.  Resampling**

Spatial sampling rate varies due to variation in the handwriting speed of the writers. In order to remove these variations, equidistant resampling is done. This operation resamples each stroke at equal intervals in space along its trajectory.

## 2.3 Segmentation

Segmentation is an operation that isolates individual characters from the handwritten text. It is done using projection profile analysis and connected component labelling. Segmentation includes the following steps:

- Line Segmentation
- Word Segmentation
- Character Segmentation

A popular technique used for line segmentation is horizontal projection profile method in which peak-valley points are identified and is used for line separation. The horizontal projection will have separated peaks and valleys if the lines are well separated. These peaks can be used to find out the boundaries between lines and so each line is extracted. Word segmentation is done by applying vertical projection profile method on the separated lines. Here the peaks and valleys are identified and words are separated by looking at the minima in the vertical profile. Finally, the characters are isolated from these words using connected component labeling. Connected components labeling groups pixels of an image into components based on pixel connectivity. This technique assigns to each connected component of a binary image a distinct label.

## 2.4 Feature Extraction

Feature extraction is the process of extracting relevant features of the characters to form feature vectors which are used by classifiers for the recognition process. The feature extraction methods for handwritten character recognition can be classified into three: Statistical, Structural and Hybrid techniques [7]. Statistical approaches use quantitative methods for extracting the features. Geometrical moments, projection histograms, direction histograms, crossing points etc. are used as features here. Structural approaches use qualitative measurements for feature extraction. These features are based on topological and geometrical properties of the character, like strokes, loops, end points, intersection points, etc. Hybrid approaches combines the features of these two techniques.

## 2.5 Classification

Classification is the final phase of character recognition, which is done by assigning labels to character images based on the features extracted. Bayesian classifier, Binary tree classifier, Nearest Neighbor classifier, Neural networks, MQDF and Support Vector Machines are some of the classifiers that are used for this purpose.

## 2.6 Post Processing

Post-processing involves steps to be taken after classification process is completed. It may include steps like representing the output in Unicode format, error correction and the disambiguation of confusing character pairs. Linguistic rules can also be applied to further improve recognition rates. These linguistic rules are specific to a language.

## 3. Malayalam Script Features

Malayalam is one among the 22 scheduled languages of India. It is the official language of Kerala and is spoken by around 35 million people in the world. Malayalam is also spoken in the Union territories of Lakshadweep and Mahe. It is one among the 4 major Dravidian languages of South India. Malayalam script is derived from the Grantha script which is an inheritor of the old Brahmi script. Malayalam has close affinity to Tamil. The Malayalam language was written in Vatteluttu, a script that had evolved from Tamil-Brahmi. It first appeared in Vazhappalli inscription.

Malayalam has the largest character set among the Indian languages. It is partially alphabetic and partially syllable based. There are about 128 characters. The basic characters can be classified as vowels/Swaraksharangal (Figure 3) and consonants/Vyanjaksharangal (Figure 4). The complete character set includes 15 vowels, 36 consonants , 5 chillu , 3 consonant signs, 9 vowel signs, anuswaram, visargam, chandrakkala and 57 conjunct consonants. It also includes 9 numerals which are rarely used.

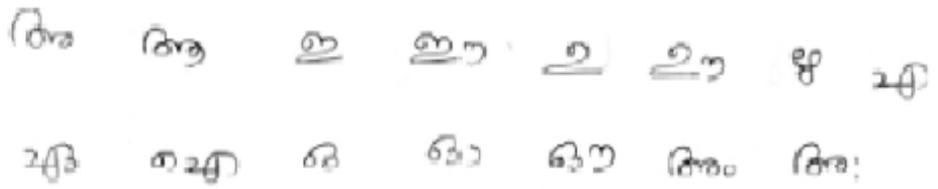

Figure 3 : Malayalam Vowels

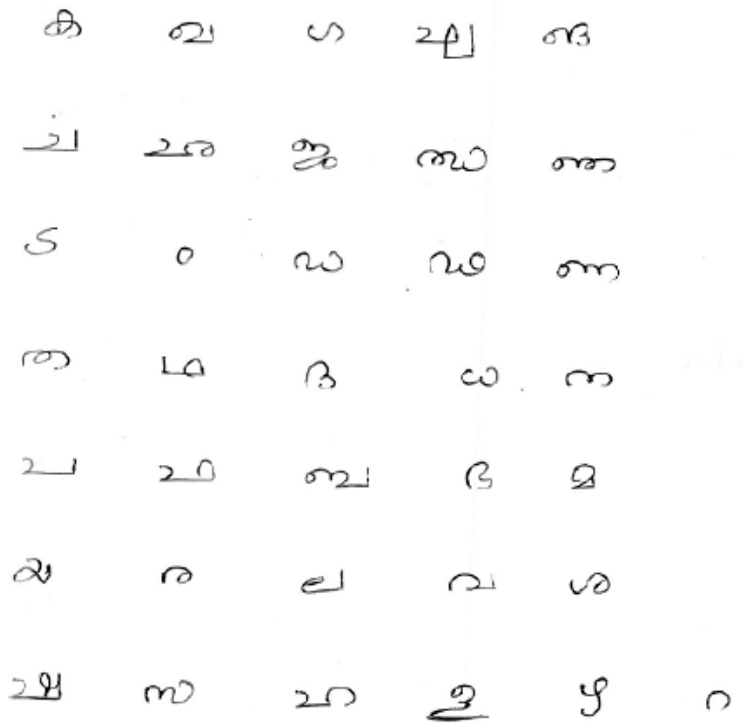

Figure 4 : Malayalam Consonants

## 4. Studies On Offline Malayalam Character Recognition

Offline character recognition is the process of recognizing handwritten text from a scanned sheet of paper. During the past few years, many works have been reported in Offline character recognition of Malayalam. The table 1 summarizes the important works that have been done so far for offline Malayalam character recognition. The first work in Malayalam OCR was reported by Lajish [8]. It used fuzzy-zoning and normalized vector distance measures for the recognition of segmented characters. The size normalized image was divided into 3x3 uniform sized zones. The normalized vector distance for each zone is computed and fuzzification is performed on these. The 9 features, thus obtained were classified using class modular neural network. The system had attained an overall accuracy of 78.87% for the 44 Malayalam handwritten characters. In [9], State space Point Distribution (SSPD) parameters obtained from gray scale based State space map (SSM) of character samples were used for classification. SSPD parameters were calculated from State-space maps with eight directional space variations. The 16 features, thus obtained were classified using class modular neural network. The system had an accuracy of 73.03%. The system had the advantage that no binarization was required and hence the difficulties in threshold selection and loss of information caused by binarization process were eliminated.

Wavelets were applied by G. Raju [10] for the recognition of isolated Malayalam characters. He used db4, a member of the Daubechie wavelet family with order 4, for decomposition into ten sub-images. The count of zero-crossing in each of the ten subbands were used for classification. From the analysis of zero crossings, 25 consonant characters that were taken in the data set could be classified into 11 sets. No preprocessing steps were done in this method. In [11], performance analysis of wavelet features using twelve different wavelet filters was done. In this work, the wavelet feature and projection profile were combined by subjecting the projection profiles to n levels of wavelet transform. A Multilayer Perceptron Network, which consists of multiple layers of computational units, interconnected in a feed-forward manner is used as the classifier. The average recognition accuracy obtained was 76.8% for selected 33 malayalam characters. The above work was also extended by adding an additional feature, aspect ratio which further improved the recognition accuracy, the average being 81.3%.

Table 1. Performance Of Offline Malayalam Techniques

| Paper | Features | Classifier | Accuracy (%) |
|---|---|---|---|
| Lajish [2007] | Fuzzy zoning,NVD | CMNN | 78.87 |
| John et al. [2007] | Wavelet | MLP | 73.8 |
| Lajish [2008] | SSPD | CMNN | 73.03 |
| Raju [2008] | Wavelet, Aspect Ratio | MLP | 81.3 |
| Chacko and Anto[2009] | Structural | SLFN | 95.18 |
| Chacko and Anto[2010] | WEF,ELM | SLFN | 95.16 |
| Rahiman et al. [2010] | HLH Intensity features | - | 88.6 |
| Rahiman et al. [2011] | VL & HL count and position | Decision tree | 91 |
| Jomy John et al. [2011] | CCH,NCCH,Image Centroid | Feedforward  NN | 72.1 |
| Jomy John et al. [2011] | Haar Wavelet Transform | SVM | 90.25 |
| Moni and Raju [2011] | RLC | MQDF | 94.18 |
| Moni and Raju [2011] | Gradient | MQDF | 95.42 |
| Vidya V et al. [2013] | Cross feature, fuzzy depth, distance & Zernike moment | PSFAM | 87.81 |

In [12], character images were modeled with projection profile. One dimensional wavelet transformation was applied on both the horizontal as well as vertical projection profile. Daubechies wavelets with filter length 4 were used for transformation. Using MLP as the classifier, a recognition accuracy of 73.8% was obtained for 33 character classes. Rahiman proposed a recognition technique based on the HLH intensity pattern of characters [13]. The characters were grouped into "ra", "pa" or special type. characters. An overall accuracy of 88.6% was achieved for the system. They have also proposed another technique for the extraction of features based on the analysis of position and count of the horizontal and vertical lines [14]. One of the main advantages of this method was that it could identify characters, even if they were written on a colored background. An accuracy of 91% was achieved in this work.

Chacko and Anto [15] proposed a method for producing smooth skeletons of Malayalam handwritten characters. Here skeleton pruning was done by contour portioning with discrete curve evolution. The overall accuracy of the system was 90.18% for 33 character classes. The recognition of handwritten Malayalam characters using the wavelet energy feature (WEF) and extreme learning machine (ELM) was proposed in [16]. ELM is an extremely fast learning algorithm for single hidden layer feed forward networks (SLFN). This algorithm learns much faster than the traditional learning algorithms for feed forward neural networks.

In [17], Jomy John et al. proposed a method based on chain code and image centroid for the recognition of Malayalam vowels. From the chain code representation of the character, a chain code histogram and Normalized chain code histogram were constructed which were used for classification process. A two layer feed forward network with scaled conjugate gradient was used for classification. An average accuracy of 72.1% was obtained for the system. The authors have also proposed another method which was based on Haar wavelet transform and used SVM as classifier [18]. Using third level decomposition, accuracy was 89.64% and using second level decomposition accuracy of 90.25% was obtained for 44 basic Malayalam characters. They have also proposed another system which uses visual image queries for retrieving similar images from database of Malayalam handwritten characters [19]. Local Binary Pattern (LBP) descriptors of the query images were extracted and were compared with the features of the images in the database for retrieving similar characters.

In [20], Bindu S Moni et al. proposed a character recognition scheme using run length count (RLC). RLC is the count of contiguous group of 1s encountered in a left to right / top to bottom scan of a character image. Modified Quadratic Discriminate function (MQDF) was used for classification. They had achieved a recognition rate of 94.18% with 72 features for 30 selected character classes. They have also proposed another character recognition system using directional features and MQDF [21]. In this method, character images were decomposed using the Fixed Meshing strategy and the twelve directional codes based on the gradient direction are extracted to form the feature vector. They have obtained a recognition rate of 95.42% using a total of 432 features for 44 malayalam characters.

Recently, Vidya V [22] proposed another approach based on Probabilistic Simplified fuzzy ARTMAP(PSFAM). Different features like cross feature, fuzzy depth, distance and Zernike moment feature are extracted for each character glyph. An accuracy of 87.81% was attained for 142 malayalam characters.

## 5. Studies On Online Malayalam Character Recognition

In online mode of writing, a user writes directly on an electronic surface with a stylus. Since temporal and spatial information about is stroke is available in online mode, it gives much better recognition rate compared to offline recognition. The table 2 summarizes the important works that have been done so far for online malayalam character recognition.

Table 2 : Performance Of Online Malayalam Techniques

| Paper | Feature | Classifier | Accuracy (%) |
|-------|---------|------------|--------------|
| Sreeraj et al. [2009] | Normalized (x,y) coordinates and context bitmap | Kohonen Networks | 88.75 |
| Sreeraj et al. [2010] | Time domain features, writing direction, curvature | KNN | 98.125 |
| Chacko and Anto[2011] | Division Point features | SLFN | 96.83 |
| Primekumar et al [2011] | Time domain features, Wavelet transform | SFAM | 97.81 |
| Indhu et al. [2012] | Structural, Directional | SFAM | 98.26 |
| Primekumar et al [2013] | Time domain features, Angular features | SVM | 97.97 |

Sreeraj et al. [23] proposed an online Malayalam character recognition technique using a combination of context bitmap and normalized (x,y) coordinates features. Classification was done using kohonen neural networks. The system reported an accuracy of 88.75% for the 64 basic characters with a recognition time of 15-32 milliseconds. They have proposed another approach for online handwritten malayalam character recognition system based on a combination of time domain features and dynamic representation of writing direction along with its curvature [24]. The various features extracted were Normalized x-y coordinates, Aspect, Curvature, Writing direction etc. Recognition was done using KNN classifier. An accuracy of 98.125% was achieved for the system. This system could provide high accuracy even with a small sample size.

In [25], OS-ELM a fast online sequential algorithm is used for single hidden layer feed forward neural networks (SLFN) with both additive and radial basis function (RBF) hidden nodes in a unified framework. They have used division point features which were generated by recursive subdivision of character images. An accuracy of 96.83% was obtained using this technique for 44 classes of characters.

In [26] a method based on wavelet Transform was proposed for the on-line recognition of Malayalam handwritten characters. In the first step six time domain features : xy co-ordinates, angular features, direction and curvature were extracted. Then the wavelet transform of these features were calculated to form the resultant feature vector in compressed form. db1and haar wavelets were used for wavelet decomposition. The system gave a maximum accuracy of 97.81% using Fuzzy ARTMAP as classifier.

The performance of the On-line Malayalam handwriting recognition system using HMM and SVM classifier is analysed in [27]. Time domain features and the angular features were used to form the feature vector. Feature extraction method based on discrete wavelet transformation is used for SVM system. The system gave maximum accuracy of 97.97% for SVM using Gaussian kernel and 95.24% for HMM. One of the major advantages of using SVM is that the training time required is very much less compared to that of HMM. Another approach based on 4 directional freeman chain coding was proposed in [28]. Starting from the point when first contact is made with the writing surface, direction in which the pen tip moves is recorded. The recognizer used was a back propagation neural network.

Indhu [29] proposed an approach using SFAM artificial neural network technique. For each character, structural and directional information is obtained. The various features extracted are the XY coordinates, Start Quadrant, End Quadrant, Horizontal & Vertical Point Density, loop, cusp, stroke length etc. Using this technique an average accuracy of 98.26% was obtained.

# 6. Conclusion & Future Works

This paper presented a detailed study on different handwritten character recognition so far developed for Malayalam language. A number of works are on progress in this area. But the problem still demands more attention. One of the major challenges encountered in this field is the lack of a benchmark database for testing. The best recognition accuracy in offline Malayalam character recognition of 95.42% was reported in [21] using gradient features and MQDF classifier. For online malayalam character recognition, the best recognition rate reported was 98.26% in [29] using structural and directional features and SFAM neural networks as classifiers. We believe that our survey will be useful for researchers who are working in the handwritten character recognition domain as we have included almost all the prominent works done in the offline and online character recognition. Most of the work in this area was done for the recognition of basic malayalam character set. Only few works have been reported for the complete character set. The future works in this area can be done for the recognition of complete Malayalam character set including conjunct characters as well. Also further works can be done for malayalam word recognition.

**Authors**

Ms. Anitha Mary M.O. Chacko is a postgraduate engineering student in Computer Science and Engineering-Information Systems at Rajagiri School of Engineering and Technology, Kochi, India. She completed her graduation in Computer Science and Engineering from P.A. Aziz College of Engineering & Technology, Trivandrum.

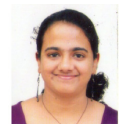

Ms. Dhanya P.M is currently working as Assistant professor in the department of Computer Science and Engineering at Rajagiri School of Engineering and Technology,Kochi,India. She took her B.Tech in Computer Science and M.Tech in Software Engineering from the Department of Computer Science CUSAT, Kochi. She is currently a Ph.d Scholar in the Department of Computer Applications ,CUSAT. Her research interests include Natural Language Processing and Data Mining.

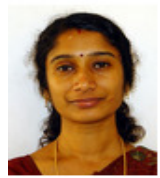